\documentclass[10pt,twocolumn,letterpaper]{article}

\usepackage{cvpr}
\usepackage{times}
\usepackage{epsfig}
\usepackage{graphicx}
\usepackage{amsmath}
\usepackage{amssymb}
\usepackage{booktabs}
\usepackage{multirow}

\usepackage[pagebackref=true,breaklinks=true,letterpaper=true,colorlinks,bookmarks=false]{hyperref}

\cvprfinalcopy 

\def\cvprPaperID{1664} 

\begin{document}

\title{Exploring Adversarial Fake Images on Face Manifold }

\author{Dongze Li$^{1, 2}$, Wei Wang$^{2 *}$, Hongxing Fan$^{1, 2}$, Jing Dong$^{2}$ \\
	$^{1}$ School of Artificial Intelligence, University of Chinese Academy of Sciences\\
	$^{2}$ Center for Research on Intelligent Perception and Computing, CASIA\\
	\texttt{lidongze2020@ia.ac.cn,hongxing.fan@cripac.ia.ac.cn}\\
	\texttt{\{wwang, jdong\}@nlpr.ia.ac.cn}
	\thanks{ Corresponding author.}
}

\maketitle
\pagestyle{empty}  
\thispagestyle{empty} 

\begin{abstract}
Images synthesized by powerful generative adversarial network (GAN) based methods have drawn moral and privacy concerns. Although image forensic models have reached great performance in detecting fake images from real ones, these models can be easily fooled with a simple adversarial attack.
But, the noise adding adversarial samples are also arousing suspicion. In this paper, instead of adding adversarial noise, we optimally search adversarial points on face manifold to generate anti-forensic fake face images. We iteratively do a gradient-descent with each small step in the latent space of a generative model, e.g. Style-GAN, to find an adversarial latent vector, which is similar to norm-based adversarial attack but in latent space. Then, the generated fake images driven by the adversarial latent vectors with the help of GANs can defeat main-stream forensic models. For examples, they make the accuracy of deepfake detection models based on Xception or EfficientNet drop from over 90\% to nearly 0\%, meanwhile maintaining high visual quality. In addition, we find manipulating noise vectors at different levels of the generator have different impacts on attack success rate, and the generated adversarial images mainly have changes on facial texture or face attributes.

\end{abstract}

\section{Introduction}

Nowadays, it is increasingly hard for human eyes to tell a real image from a fake one with the rapid improvement of image generation techniques. Manipulated or generated fake images may draw social and privacy concern if being abused by malicious attackers. An attacker may register an account with photos belonging to a non-existent person or swap one person’s face to another, thus causing privacy and security issues. Image forensic models are designed to clarify those images from real ones, and have gained considerable performance on several benchmarks and datasets \cite{li2019celeb,rossler2019faceforensics++}. However, a smarter attacker may attempt to generate images which can bypass those detectors while keeping high visual quality. These images may escape the detection procedure and spread in social media. In order to combat the generation and spread of undetectable ``deep fakes", it is necessary for image forensics researchers themselves to develop and study anti-forensic operations\cite{2011Anti} to improve the robustness and generalization ability of the existed forensic models. 

\begin{figure}[t]
\begin{center}
   \includegraphics[width=0.8\linewidth]{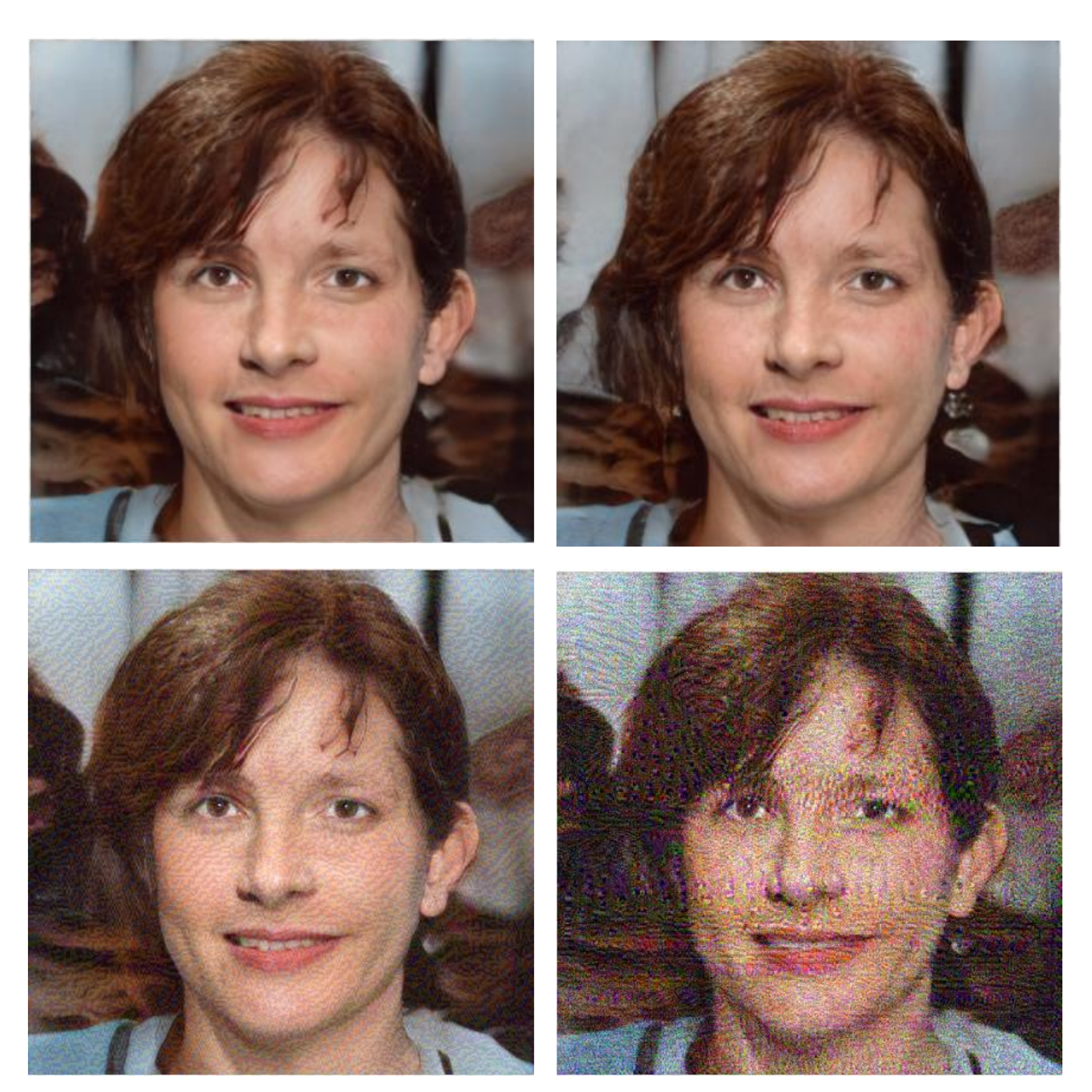}
\end{center}
   \caption{Adversarial images generated by different methods. Upper left is the original Style-GAN-generated image. Upper right is the image generated by our method. Lower left and Lower right are adversarial images generated by FGSM\cite{goodfellow2014explaining} and PGD\cite{madry2017towards} $L_{inf}$ norm-based attack respectively under the same perturbation level. Although all these images can bypass the target forensic model, images generated by our method are more invisible to human eyes.}
\label{fig:1}
\end{figure}

In this paper, we propose to efficiently generate adversarial high visual quality fake images to fool forensic detectors. By adversarially exploring on the manifold of the recent powerful generative model Style-GAN\cite{karras2019style}, we can therefore generate the adversarial fake face images that fool forensic models. Though StyleGAN is capable of generating high-resolution images with various styles and stochastic details, the generated fake images are easily detected by models based on Xception or EfficientNet with accuracy of over 90\%. But with intentionally iteratively searching these adversarial vectors in its latent space with a gradient descent manner, we successfully screen out fakes images that will be detected by forensic models as real ones.

Although one can exploit existing adversarial attack methods \cite{goodfellow2014explaining,madry2017towards} to deceive a forensic model, it may hold visible perturbations brought by the optimization process in image space, which make it detectable by human eyes or specially designed detectors \cite{2017Towards}. Recently proposed unrestricted adversarial attack methods \cite{song2018constructing,wang2019gan,kakizaki2019adversarial} could generate adversarial images with less suspicious visual artifacts by training GAN models, but they mainly focus on defeating classification or recognition tasks.

Our method has superiority in the following aspects: First, because we do modifications on the manifold, we don’t have to care much about the image pixel constraint, which makes a higher updating strength possible. Second, unlike norm-based attack which leave visible artifacts onto the image, our method can generate the same image without obvious artifacts, see in Figure~\ref{fig:1}.

Our contribution are as follows:

1. We propose a novel method of generating adversarial anti-forensic images via exploring Style-GAN's manifold. Images generated by our method can successfully bypass two image forensic models, Xception~\cite{chollet2017xception} and EfficientNet~\cite{tan2019efficientnet}. indicating the demand for more robust forensic models. 

2. We compare our method with nowadays widely-applied norm-based adversarial attacks and show that the proposed method can achieve the same attack success rate while introducing less visible perturbation, making it harder for our adversarial image to be detected by human eyes.

3. We conduct our attack in different ensemble ways and have shown our adversarial images can transfer between different forensic models, causing a threat even in the situation where the architecture of forensic models is unknown to the attacker. 

4. We show some interesting effects between the adversarial strength, the level of input noise vector and the attributes of our generated images, which are worthy of investigation in future.
\section{Related Work}
\subsection{Forged Image Generation and Detection}


\textbf{Forged Image Generation}.
Generating forged images manually with image-editing tools can be time-consuming and may be easily detected  by  both  human  eyes  and  other forensic  methods. Recent deep-learning methods lower the threshold of synthesizing these fake images, making it possible to yield a large volume in a short period. Nowadays forged images have drawn more attention since detection on them are much harder then before, and have caused serious privacy and security issues, in which fake facial images account for a large proportion. There exists several methods to synthesize forge facial images\cite{tolosana2020deepfakes}: Entire face synthesis \cite{karras2019style,mirza2014conditional,karras2017progressive}, which our method belongs to, uses GANs or other generative models to generate a fake image; face identity swap\cite{korshunova2017fast,thies2016face2face,nirkin2019fsgan}, which swap one person's face to another one; face attributes editing \cite{he2019attgan,zhang2018generative},which manipulates face attributes with image editing software or deep learning models and face expression manipulating \cite{garrido2014automatic,thies2016face2face,nirkin2019fsgan},  which transplant a target's facial expression to a source person.

GAN framework was first introduced by Goodfellow \etal~\cite{goodfellow2014generative} and has been widely applied to a series of unsupervised and semi-supervised image generation tasks . The main idea of a GAN is to fit the data distribution with an adversarial game procedure, where a generator tries to deceive a discriminator which aims to tell an image is from the real data distribution or not. GAN framework can also realize cross-domain translations \cite{isola2017image,zhu2017unpaired} with additional information and carefully designed loss functions. GANs have shown awesome results in face generation \cite{mirza2014conditional,karras2017progressive,karras2019style},and have drawn broad awareness for the vital role face images play nowadays. We choose Style-GAN~\cite{karras2019style} pretrained on FFHQ dataset for the diversity of its latent space and its ability to generate high resolution images with various styles and fine grained details. With a strong generator, we can generate more realistic images to bypass detectors as well as human eyes.\\

\textbf{Image Forensics}. Traditional Image forensic approaches are usually based on specific artifacts left by a certain forge method, which lack versatility and is fragile to the change of forge method and data distribution. Recent machine-learning and deep-learning methods are capable to handle more complex forge approaches. 
Zhang \etal~\cite{zhang2017automated} used SVMs and Random Forests to classify forged facial image, which is the first work to use machine-learning method on image forensics. A two-stream network was proposed to by Zhou\etal~\cite{zhou2017two} for face manipulation detection.
MesoNet proposed by Yamagishi\etal~\cite{afchar2018mesonet} uses a network with low layer numbers, focusing on the mesoscopic properties of images, to detect manipulated images.  
Rossler \etal~\cite{rossler2019faceforensics++} shows that a Xception model outperforms than other model on the forged image classification task. In practice, EfficientNet \cite{tan2019efficientnet} also show good performance.

Recently, a Face X-Ray method \cite{li2020face} which focus on the detection of image blending artifact has been proposed and have shown good performance on identity-swapping.  

For the simplicity, we focus on bypassing deep learning forensic models which are aimed to detect whether an image derives from a GAN or not. Images generated by our method is able to escape the detection of two selected forensic models, Xception \cite{chollet2017xception} and EfficientNet \cite{tan2019efficientnet}. 


\subsection{Adversarial Examples}
Adversarial Examples are images crafted with small modifications to fool a target classifier. Given a classifier f and an clean image $X$ as well as its ground truth label $y$ which belong to a label set $S$,where $f(x)=y$ for an well-behaved classifier. The goal of an adversarial attack is to get a modified image $X'$ and make the threat model predict $f(X')=y’$ while $y' \in S$ and $y' \neq y$ . The type of the attack can be categorized as non-targeted attack and targeted attack based on whether its goal is to make the target model classify the adversarial image as any $y' \neq y$ or a specified $y'$. Based on how much information about the target model is known, attack method can be divided into white-box attack, in which model weights and architectures are accessible, and black-box attack, where attackers hold limited knowledge about the model. Besides, adversarial examples have shown transferability \cite{papernot2016transferability,liu2016delving}, that is, adversarial examples crafted on one model may also fool other models although their architectures are different.
The norm-based attack methods require the distance between the crafted adversarial image $X'$ and the original image $X$ should satisfy the p-norm constraint $\left \| X'-X \right \|_{p} < \epsilon $. Series of work have followed this protocol \cite{dong2018boosting,madry2017towards,rony2019decoupling,wu2020boosting} to improve the attack strength and transferability. Several ways to resist adversarial examples have also been proposed, such as adversarial training\cite{goodfellow2014explaining,madry2017towards}, gradient distillation\cite{papernot2016distillation}, high level guided denoise\cite{liao2018defense} and so on. Our work has taken advantage of several attack methods below but is free from the norm constraint. \\

\textbf{FGSM}. 
FGSM \cite{goodfellow2014explaining} is an one-step attack method which add perturbations onto the original images, hoping to maximize the loss function $J(X',y)$, while $J$ is usually the cross-entropy loss, FGSM is formally defined by 
\begin{equation}
X'=X+\epsilon \cdot sign(\nabla_{X} J(X,y)),
\end{equation}
where $\epsilon$ denotes the max perturbation scale. We have deployed FGSM attack onto our generated images, trying to fool the forensic model as a controlled experiment. \\

\textbf{PGD}. 
Madry et.al \cite{madry2017towards} deploy a strong iterative attack method called Projected Gradient Descent. in each step the perturbation is projected to a $\epsilon -$ based ball. Defined by 
\begin{equation}
X_{t+1}=X_{t}+\alpha \cdot sign(\nabla_{X_t}J(X_{t},y)),
\end{equation}
in which $X_{t}$ denotes the attack image in t-th iteration. 

In our work,we use a similar way to update our input vector to search on the generator's manifold, we also deployed $L_{inf}$ and $L_{2}$ PGD attack onto the Style-GAN generated image to compare with on method on visual quaility and attack success rate.\\

\textbf{Unrestricted Adversarial Examples}. 
The traditional adversarial perturbations are constrained by norm-bounds, where unrestricted adversarial examples are free from. 
To generate an unrestricted adversarial image, one can apply various modifications to the original image,such as spatial transformations \cite{xiao2018spatially}, rotating \cite{engstrom2017rotation},attribute editing \cite{qiu2019semanticadv}, translations\cite{kakizaki2019adversarial}, or even construct an image from scratch \cite{song2018constructing}, as soon as the synthesized image still belongs to its previous class, which are usually judged by an auxiliary classifier\cite{song2018constructing}.A recent work\cite{poursaeed2019fine} also takes advantage of Style-GAN, generating unrestricted adversarial images via modifying Style-GAN's style vectors and noise inputs, attacking image classification models on ImageNet\cite{deng2009imagenet}, CelebA\cite{liu2015faceattributes} and Lsun\cite{yu2015lsun}, while we are aimed at deceiving forensic models and focus on face images generated by Style-GAN for the unique role face images plays in image forensic applications. 

\begin{figure*}
\begin{center}
\includegraphics[width=1.0\linewidth]{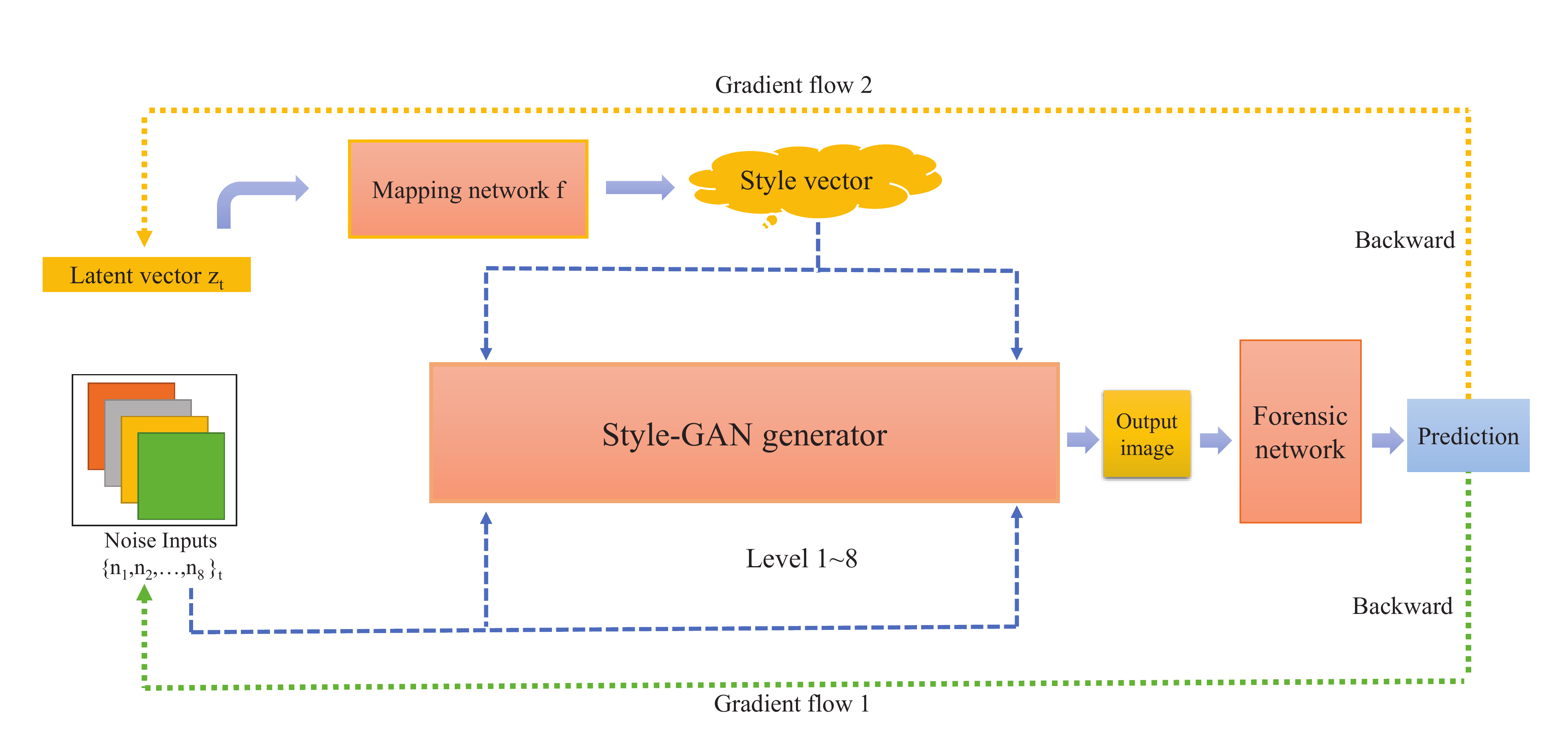}
\end{center}
   \caption{The overall pipeline of our method. We perform gradient descent on the latent vector and noise inputs of Style-GAN, respectively or together, maximizing loss function of the target forensic model(s). }
\label{fig:3}
\end{figure*}

\section{Method}

Although highly realistic images can be generated by Style-GAN, they are easily to be detected by a forensic model. In this paper, we try to generate fake images which can fool forensic detectors without quality degradation. We do manipulations on Style-GAN's input latent vector $z$ or noise vector $n$ in the neighborhood of the given vector to find adversarial images on the face manifold. similar to work \cite{poursaeed2019fine}, while keeping the original architecture of Style-GAN unchanged. 

\subsection{Style-GAN}
In our task, what we need is the original pretrained Style-GAN and its weight. Style-GAN architecture consists of a 8-layer linear mapping network $f$ which maps a latent code $z$ to an intermediate style space to get style vector $w$, and a generator $g$ whose each layer takes style vectors and random noise inputs as input and generates image progressively. High-level patterns are determined by style vectors and stochastic details are controlled by noise inputs, respectively. 

\subsection{Anti-Forensics Fake Image Generation}
In this work, we aim at generating our image adversarial once for all avoiding extra adding perturbations operation. As the iteration process goes further, the adversary of our generate images becomes stronger. Our search method focus on finding the right direction which makes the generated images have the ability to escape the forensic models. Motivated by the traditional norm-based iterative adversarial attack, we apply gradient descent on the noise and latent vector of Style-GAN, updating those vectors towards the direction which maximizes the loss of the forensic model's prediction. Each time after updating, we validate whether the forensic detector predicts the fake image as real. Because forged image detection task is a two classification problem, a non-target method will suffice. We bypass the target detection models successfully if they misclassify our generated images as real. The whole pipeline of our method is shown in Figure~\ref{fig:3}. 

In the basic attack setting, we minimize the binary cross-entropy loss of the final prediction of a single forensic model, while in the ensemble setting, the predictions of the model are combined in different strategies, to achieve higher attack success rates on both models.

\subsection{Adversarially Searching on Face Manifold}
We perform gradient descent to the latent vector $z$ before the mapping network $f$ and the noise inputs $n_1,n_2...n_8$ inserted at each level of the generator's upsampling layer. Each time, we use the one step gradient sign of the loss function to update our latent vectors with a fixed step size hyperparameter $\epsilon_1$. 
The modification formula can then be described as 
\begin{equation}
z_{t+1}=z_t+\epsilon_{1} \cdot sign(\nabla_{z_t}(J(g(z_{t},n_{t}),y))),
\end{equation}
where $z$ denotes the input latent vector and $n$ represents the noise injected into each layer, $y$ is the ground truth label of the generated images and is always set to fake, $g$ is our pretrained Style-GAN generator. It is similar for each level's noise inputs. The iteration formula for noise inputs can be written as
\begin{equation}
n_{t+1}=n_t+\epsilon_{2} \cdot sign(\nabla_{n_t}(J(g(z_{t},n_{t}),y))).
\end{equation}
The reason why we don't update the style vector after the mapping network $f$ directly is that we want to keep the integrity of the whole Style-GAN architecture.

%
Updating on noise inputs $n_1,n_2,...,n_8$ and latent vector $z$ can be carried out separately or together. Each level of the generator network gets style vectors derived from the same $z$, at the first few steps of the iteration process, the images remain the same as they were. When iteration number increases, the texture of the resultant images become much deeper, see in Figure~\ref{fig:6}. Modification on $n_1,n_2,...n_8$ will have more interesting effects on generated image for it will gradually change the appearance of the output image after more iteration steps and will bring different adversarial strength. Synthesized images have the ability to bypass the detector with ignorable changes after the first 1-2 steps, while after 3-5 steps, some attributes of the generated person's face have obviously changed. We also found that different update step size and iteration times have impact on the generate images visual quality. So we have to select feasible step sizes and iteration times if we want to keep the certain appearance of a person. The fact also means we might be able to deploy successful attacks with some of the image attributes modified, while others stay constant. 

Recently, \cite{poursaeed2019fine} also conducted adversarial attack through Style-GAN, which seems to be similar to our method. We clarify that our method is different from theirs in the following aspects. First of all, \cite{poursaeed2019fine} aims at attacking more general tasks, such as image classification and instance segmentation. while our work concentrates more on a specific topic, Deepfake detection. Secondly, \cite{poursaeed2019fine} mainly aims to disable a classification task, so their attack will need an input image (also called reference image) which exists in the real world, while our goal is to generate fake images that can evade forensic detection totally from scratch.
Lastly, \cite{poursaeed2019fine} only reports performance on white box attack scenario, while the proposed method additionally shows promising performance under a black box attack protocol, which will be shown in section 4.2. 

\subsection{Gradients Ensemble}
To increase the attack success rate of adversarial images generated by our method, we propose to conduct ensemble attack. We organize our ensemble attack method in three different manners: an alternatively attack manner, ensemble in loss and ensemble in network predicted logits. The success of the ensemble attack suggests our method can be used to bypass a set of forensic models as far as we know their architecture. To bypass a single model we need to search through a direction which maximizes the target model's predict error, while ensembling can constrain our searching towards the direction where we can find the images which make the prediction of both models deteriorate. Gradients ensemble can effectively avoid searching into local maximums.

\section{Experiment}
\subsection{Basic Setting}

Our method generates images totally from scratch. We use Style-GAN generator pretrained on FFHQ dataset as our generator. Random latent vectors and noise inputs are sampled from a standard Gaussian distribution. 
As for the forensic models, we use EfficientNet-B3 
\cite{tan2019efficientnet}and Xception \cite{chollet2017xception} network for training, both of which are loaded with pre-trained weights on the imagenet dataset. The training data set is composed of images from FFHQ Dataset and StyleGAN generated images, we use the first 50,000 images in both image sets as our training set, the 50,000-60,000th picture as the validation set, and the 60,000-70,000th picture as the test set. The size of the pictures in the dataset is $1024\times1024$. We resize those images to $299\times299$ and $300\times300$, and feed them into the EfficientNet-B3 and Xception networks respectively. We use Adam \cite{kingma2014adam} as the optimizer and set the learning rate to 0.0002, weight decay to 0.001, and epoch to 3. In each epoch, we verified accuracy of our model for 5 times, and we choose the model with the best performance in the verification process as our final model and evaluate them on the testset.

To construct our attack dataset, for each strategy, we follow the same process: we first generate 5,000 images from scratch as our fake GAN-generated images, labelled as 1, and select another 5,000 real image from FFHQ dataset as real images, labelled as 0. Image size is $1024\times1024$ after generation process and is resized then normalized to $[0,1]$ before being fed into forensic models. We test the accuracy of the detectors on those real images and fake images without attack as models accuracy on clean images. Average accuracy of each model has reached over 90\%. Next, we apply our attack on those generated fake images and test the accuracy of the detector on the attacked images.

\subsection{White-box Attack}

In our basic white-box attack setting, we conduct three ways to attack. Noise method means we perform gradient decent on all noise inputs with the latent vector $z$ fixed, while latent method means we only update the latent vector $z$ with the noise inputs fixed. The third method, noise and latent means simultaneously updating noise inputs $n_1,n_2...n_8$ and latent vector $z$. As our threat model, Xception and EfficientNet are attacked separately. The hyperparameter $\epsilon_{1}$ and $\epsilon_{2}$ are set to 0.004 and 0.05, respectively. Although most of the attack images are able to bypass the target forensic models after 1-2 steps, we continue updating on these vectors till iteration times reached 10 for we want to see how the distortion the updating process bring to the output images will be. 

As baseline experiments, we develop a FGSM attack under $L_{inf}$ constraint and two PGD attacks under $L_{inf}$ and $L_2$ norm constraint on the target forensic models. For $L_{inf}$ and $L_{2}$ PGD attack, the iteration step size $\epsilon$ is set to 0.01, and will last 40 times. All pixel values are normalized to [0,1] in our experiment, and will be clipped after each iteration to avoid invalid outputs. and the allowed maximum perturbation size is 0.3 in all norm-based attacks. 

Our basic attack success rate on two forensic models, comparing with the PGD attack, are shown in Table~\ref{tab:1}. In our attack setting, before attack, the accuracy on both real images and fake images is over $90\% $, while after our attack, the accuracy have dropped to less than $1\%$. Baseline methods also perform well in degrading the prediction of both forensic models, except for PGD $L_2$ attack, which only decrease the accuracy of the model by about $30\% $. We analyze that it is because of $L_2$ adversarial attack takes the whole image's information in consideration and thus requires much larger perturbation scale. While most methods shown in Table~\ref{tab:1} can successfully deceive the model, our method is better in the visual quality of generated images. Images generated by our method, $L_{inf}$ PGD attack and FGSM are shown in Figure~\ref{fig:1} . PGD $L_2$ attack is not in our consideration due to its low attack success rate.

\begin{table}[]
\begin{center}
\begin{tabular}{l|llll}
\cline{1-3}
\multicolumn{1}{c|}{\multirow{2}{*}{Method}} & \multicolumn{2}{c}{Model} &  &  \\ \cline{2-3}
\multicolumn{1}{c|}{}                        & EfficientNet  & Xception  &  &  \\ \cline{1-3}
Clean image                                  & 97\%         & 93\%     &  &  \\
PGD $L_{inf} (\epsilon=0.3)$                                     & \textbf{0\%}             & 5\%      &  &  \\
PGD $L_2 (\epsilon=0.3)$                                       & 60\%         & 63\%      &  &  \\
FGSM $L_{inf}$ $(\epsilon=0.3)$                                         & 13\%          & 5\%      &  &  \\ 
Noise(ours)                             & \textbf{0\%}             & \textbf{0\%}         &  &  \\
Latent(ours)                            & \textbf{0\%}             & \textbf{0\%}         &  &  \\
Noise and latent(ours)                         & \textbf{0\%}             & \textbf{0\%}         &  &  \\\cline{1-3}
\end{tabular}
\end{center}
\caption{Accuracy different models perform on our method and other adversarial attack method, PGD $L_2$, PGD $L_{inf}$ and FGSM attack. our method has the same ability to bypass the forensic detectors as norm-based adversarial attack, PGD $L_{inf}$ and has better adversarial strength than FGSM and PGD $L_2$ attack. PGD $L_2$ attack shows poor performance on both models because of the limited perturbation scale.}
\label{tab:1}
\end{table}

\subsection{Black-Box Attack}
In this subsection, we report the performance of our generated images in a black-box setting. We  explore the black-box transferability of our generated adversarial image from one single forensic model to another. First we set our attack strategy to be the latent noise method as mentioned above, and fix the iteration step to 3. Under this situation, variations in images attributes is relatively small, meanwhile result images have enough adversarial strength. For the two forensic models, we generate 5000 adversarial images, denoted as $Img_{e}$ and $Img_{x}$. We then test the accuracy Xception model reaches on adversarial images generated on EfficientNet and vice versa.   

We observe that images generated on EfficientNet by our method shows good black-box transferability, decreasing the accuracy of Xception from $93\%$ to $50\%$ by nearly a half. While images generated on Xception shows little transferability, the accuracy of EfficientNet is $97\%$ on clean GAN-generated images before attack, while it is $90\%$ on our adversarial images. Images generated by $L_{inf}$ PGD attack, $L_2$ PGD attack and FGSM are also tested in the same way, and similar result was found. The model accuracy of black-box attacks can be found in Table~\ref{tab:2}. From the chart we can find  transferability of norm-based attacks are better than our method. We speculate the reason for the fact is 1: Our method are generated from the specific Style-GAN's manifold and they still have artifacts(although invisible to human eyes) that can be caught by detectors and 2: norm based attack have less overfitting than our method on the target model. 

\begin{table}[]
\begin{center}
\begin{tabular}{l|l|lll}
\cline{1-4}
\multicolumn{1}{c|}{\multirow{2}{*}{Target Model}} & \multicolumn{1}{c|}{\multirow{2}{*}{Method}} & \multicolumn{2}{c}{Test Model}               &  \\ \cline{3-4}
\multicolumn{1}{c|}{}                              & \multicolumn{1}{c|}{}                        & \multicolumn{1}{l|}{EfficientNet} & Xception &  \\ \cline{1-4}
\multirow{4}{*}{EfficientNet}                      & FGSM $L_{inf}$                                    & \multicolumn{1}{l|}{\textbf{0\%}}          & \textbf{0\%}      &  \\
                                                   & PGD $L_{inf}$                                     & \multicolumn{1}{l|}{\textbf{0\%}}          & \textbf{0\%}      &  \\
                                                   & PGD $L_2$                                       & \multicolumn{1}{l|}{60\%}         & 81\%     &  \\
                                                   & Ours                                         & \multicolumn{1}{l|}{\textbf{0\%}}          & 50\%      &  \\ \cline{1-4}
\multirow{4}{*}{Xception}                          & FGSM $L_{inf}$                                    & \multicolumn{1}{l|}{\textbf{13\%}}         & 5\%      &  \\
                                                   & PGD $L_{inf}$                                     & \multicolumn{1}{l|}{86\%}         & \textbf{0\%}      &  \\
                                                   & PGD $L_2$                                       & \multicolumn{1}{l|}{95\%}         & 63\%     &  \\
                                                   & Ours                                         & \multicolumn{1}{l|}{90\%}          & \textbf{0\%}      &  \\ \cline{1-4}
\end{tabular}
\end{center}
\caption{Model accuracy under black-box setting. Adversarial examples generated from norm-based attacks (except $L_2$ PGD attack shows better transferability. while our method has also shown transferability to some extent. }
\label{tab:2}
\end{table}

\begin{figure}
\begin{center}
\includegraphics[width=1.0\linewidth]{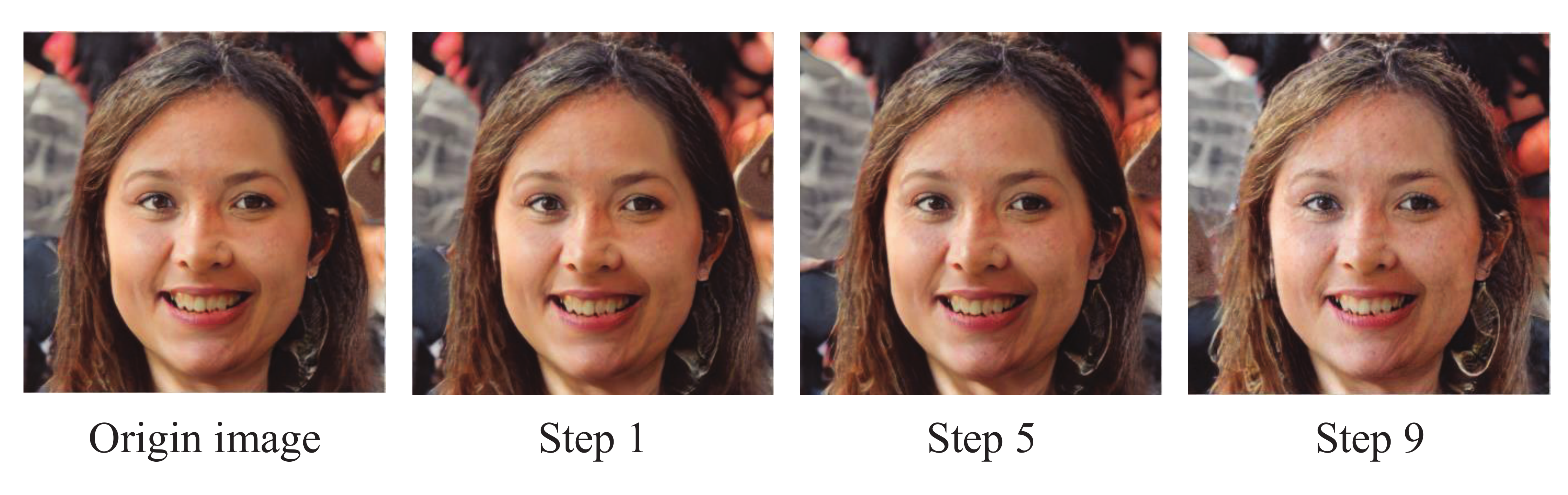}
\end{center}
   \caption{Images generated though solely updating latent vector z at different steps. $\epsilon_{1}$ is set to 0.004. While successfully deceiving the target model,
   our adversarial images are the same as the original image except for some nearly invisible textures under proper iteration times and step size.}
\label{fig:6}
\end{figure}

\subsection{Ensemble Attack}
We have found that attack between different models which relies on transferability is kind of fragile and lacks robustness. To improve the performance, we conducted our experiment in an ensemble way. we have tried three ensemble methods: ensemble in loss, ensemble in logits, and a an alternative attack manner. Ensemble in loss means loss function of the two forensic models are added together after each calculation, while in ensembling in logits setting, we fuse output logits of image forensic models to get the final cross-entropy loss function, which is then used to get the gradient with a backward pass, and in this setup,the weight of the EfficientNet and Xception are both 0.5 for we observed their gradients are in the similar scale. In the alternative attack scenario, we alternatively carry out gradient descent updating process according to one model in the ensemble models in each step. Other settings are same as the base attack strategy. Result of model accuracy on different ensemble attack method are shown in Table~\ref{tab:3}. Result on Xception and EfficientNet model show all the ensemble attack method are able to deceive the target forensic model with a high confidence. The success of different ways of ensemble attack suggests we can improve anti-forensic ability of our generated images by expand our target model set.  

\begin{table}[h]
\begin{center}
\begin{tabular}{lclll}
\cline{1-3}
\multicolumn{1}{l|}{\multirow{2}{*}{Ensemble Method}}    & \multicolumn{2}{c}{Model}                   &  &  \\ \cline{2-3}
\multicolumn{1}{l|}{}                   & \multicolumn{1}{l}{EfficientNet} & Xception &  &  \\ \cline{1-3}
\multicolumn{1}{l|}{Ensemble in loss}   & \multicolumn{1}{l}{0.5\%}        & \textbf{0\%}        &  &  \\ 
\multicolumn{1}{l|}{Ensemble in logits} & \multicolumn{1}{l}{\textbf{0.1\%}}        & \textbf{0\%}        &  &  \\
\multicolumn{1}{l|}{Alternative attack} & \multicolumn{1}{l}{0.4\%}        & \textbf{0\%}        &  &  \\ \cline{1-3}
                                       
\end{tabular}
\end{center}
\caption{Model accuracy under different ensemble attacks. Images generated from all the attack method can bypass the model successfully}
\label{tab:3}
\end{table}

\subsection {Impact of Noise Inputs on Generated Images}

\textbf{Noise Level}. Updating on the latent vector $z$ and noise inputs $n_1,n_2,...n_8$ can both achieve the goal of deceiving the target model. When iteration times are small enough, the result image have little change comparing with the original one in both settings. However, When the iteration times goes larger, updating on $z$ only brings irregular textures to the output image, making the attack more fragile facing with human eyes,see in Figure~\ref{fig:6}. while updating on noise vector shows more interesting results, controlling both the iteration step scale and updating noise injected in different level of the generator can affect both the attack success rate and the character's appearance of the generated image.

Concretely, changing the noise injected in lower level of the generator may affect the output character's appearance such like gender, beard, hair cuts and so on, while changing higher level noise may affect more general features of the images like color or textures. Updating on a certain level also leads to different attack strength. 

\begin{figure}
\begin{center}
\includegraphics[width=1.0\linewidth]{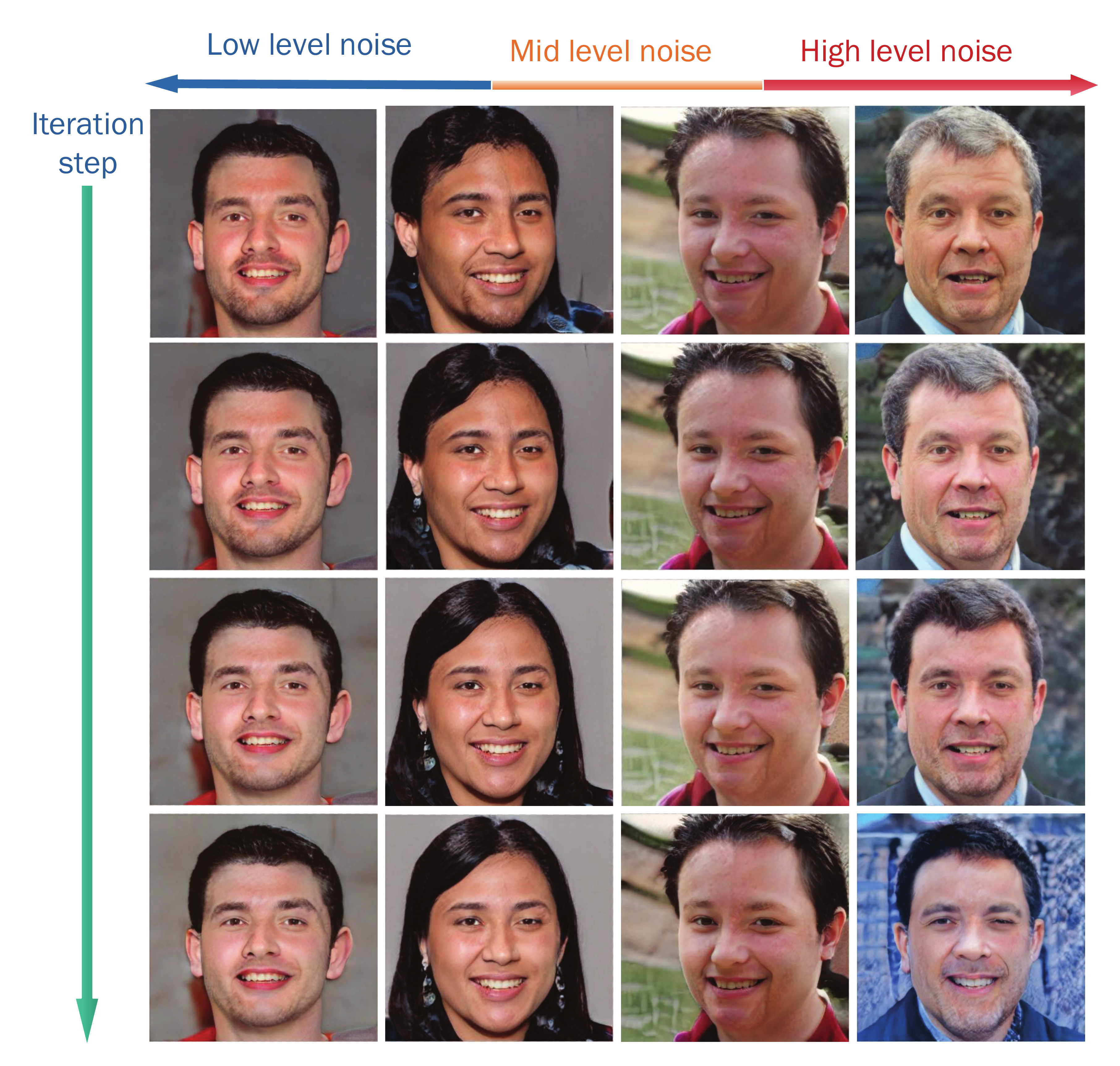}
\end{center}
   \caption{Images generated though solely updating noise vector at a specific level of the generator at different iteration steps. Updating low-level noise inputs results in more obvious attribute shifts.}
\label{fig:5}
\end{figure}

We fix the latent vector z, and conduct attack by respectively updating noise vector from level 1 to level 8. We choose images generated in iteration step 3 and set a proper step length $\epsilon_{2}=0.05$ for it won't be too large to make the result images suffer from serious distortions. We have found that while the iteration steps and the latent vector $z$ is fixed, the higher level noise vector impact the model prediction most, in other words, updating high level noise have strongest adversarial impact. Figure~\ref{fig:5} shows our generate images under this setting and attack success rates of updating different level of noises are shown in Table~\ref{tab:4}. 

\begin{figure}
\begin{center}
\includegraphics[width=1.0\linewidth]{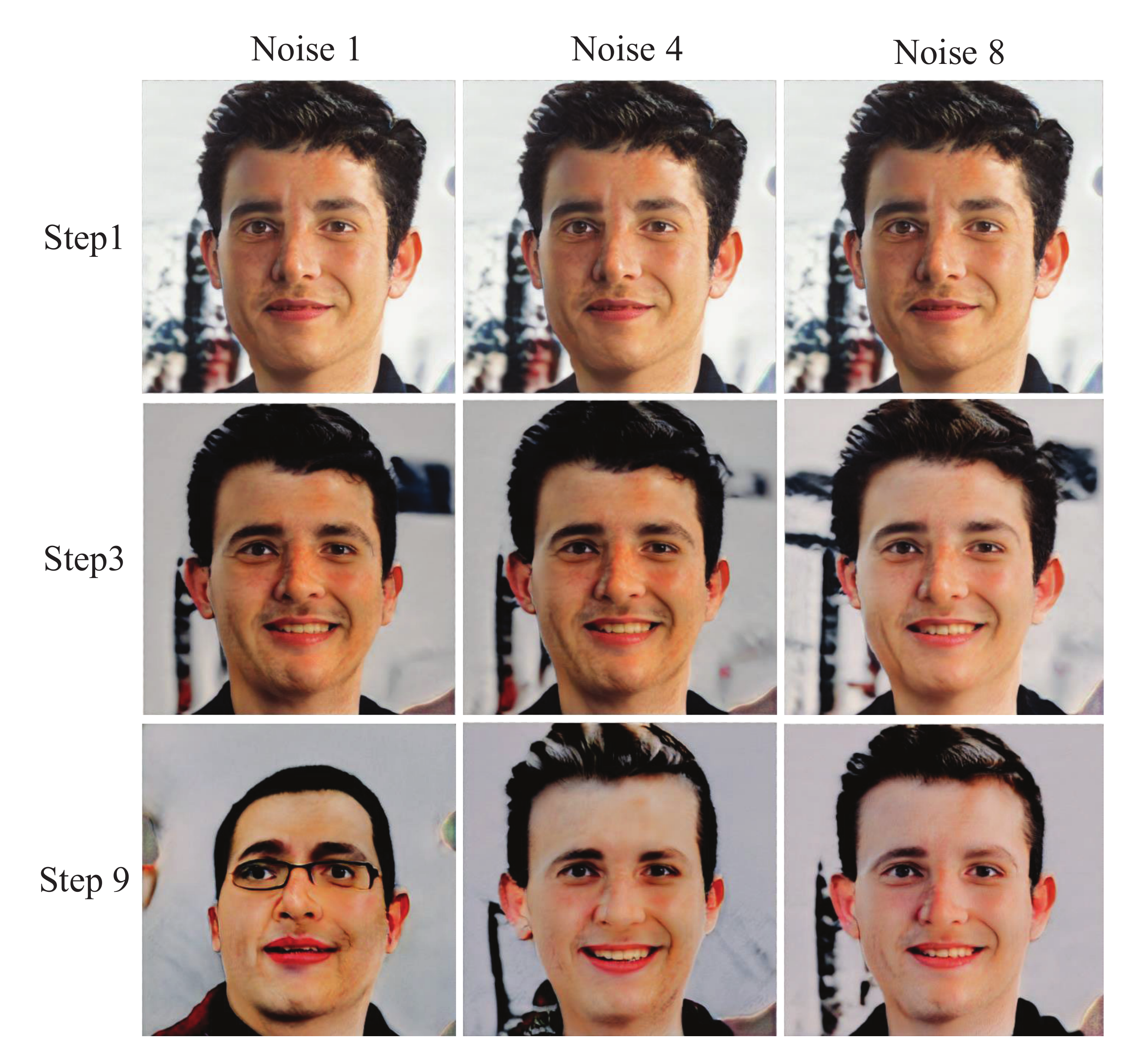}
\end{center}
   \caption{Images for a single person generated with a relatively large stride at different iteration step. After 3 steps, resultant images are still similar to their counterpart. While after 9 steps, distortion has become large enough to be noticed by human eyes.}
\label{fig:7}
\end{figure}

\begin{figure}
\begin{center}
\includegraphics[width=1.0\linewidth]{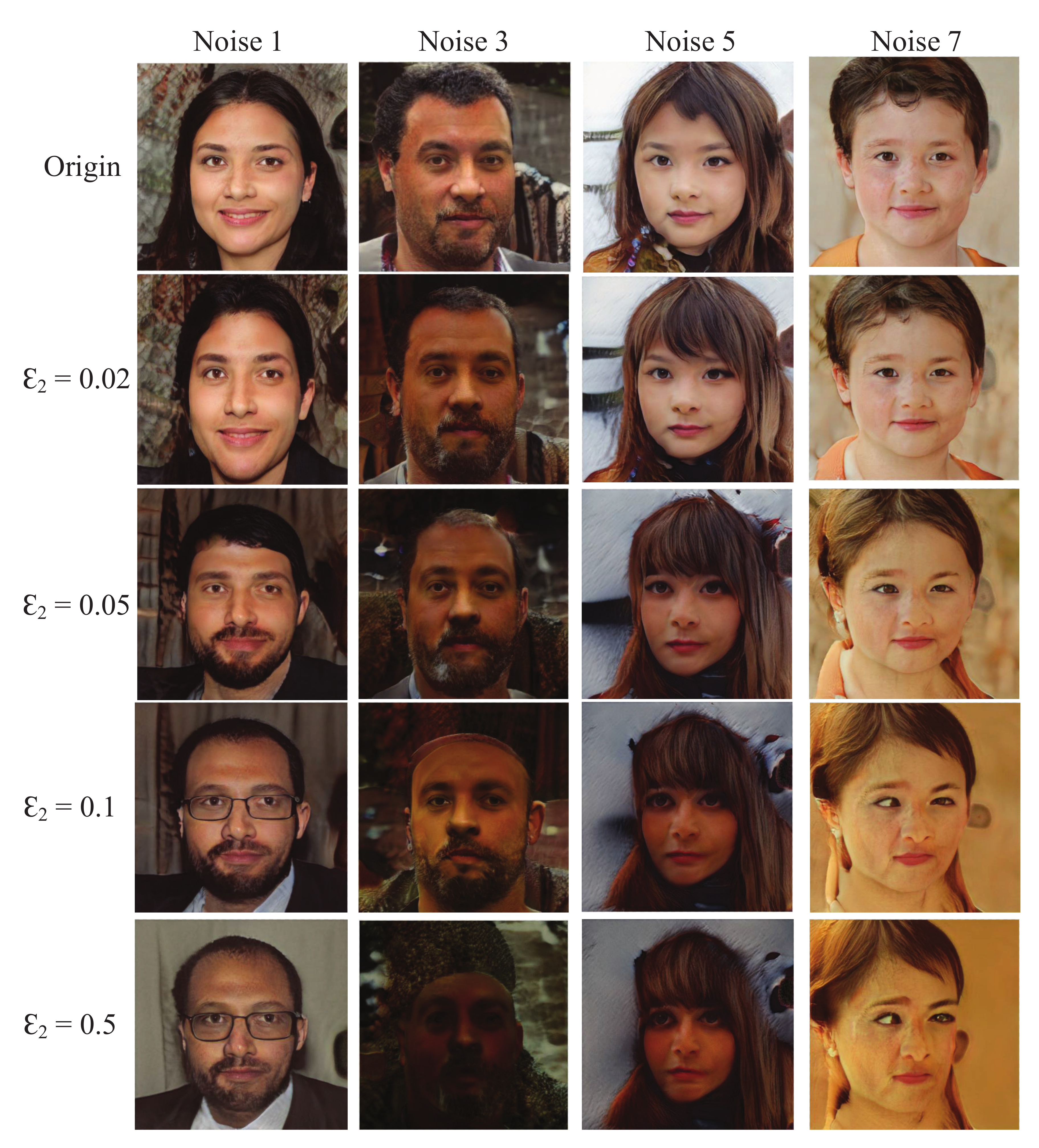}
\end{center}
   \caption{Images generated though with different stride at iteration step 3. Overlarge stride will lead to poor image quality.}
\label{fig:8}
\end{figure}

\begin{table}[]
\begin{center}
\begin{tabular}{l|clll}
\cline{1-3}
\multicolumn{1}{l|}{\multirow{2}{*}{Attack level}}            & \multicolumn{2}{c}{Model} &  &  \\ \cline{2-3}
 & EfficientNet    & Xception   &  &  \\ \cline{1-3}
0 (no attack)  & 97\%         & 93\%       &  &  \\ 
1            & 99\%         & 87\%       &  &  \\
2            & 98\%         & 88\%       &  &  \\
3            & 98\%         & 85\%       &  &  \\
4            & 75\%         & 57\%       &  &  \\
5            & 35\%         & 23\%       &  &  \\
6            & \textbf{1\%}         & \textbf{0\%}          &  &  \\
7            & \textbf{1\%}         & \textbf{0\%}          &  &  \\
8            & \textbf{1\%}         & \textbf{0\%}          &  &  \\ \cline{1-3}
\end{tabular}
\end{center}
\caption{Model accuracy on adversarial images generate with only one level‘s noise vector taking part in the updating process. Updating higher level noise vectors results in stronger adversary.}
\label{tab:4}
\end{table}

\textbf{Iterations \& Step Size}. We also found when updating step on our noise vector become larger, Changing in lower level noise inputs can also result in variations on some face attributes. Images generated through updating on different noise level with different iteration steps are shown in Figure~\ref{fig:7} (with fixed $\epsilon_{2}=0.05$). More iteration steps bring larger image diversity.  While proper modifications on noise inputs yield feasible results, excessively updating or overlarge step-size can cause serious distortions. We show images of a single person under different iteration steps with $\epsilon_{2}=0.1$ in figure and images with different iteration step size while step=3 in Figure~\ref{fig:8}.

While the clear relationship among adversarial strength, image attributes and noise level is still not clear. Further research may make us have the ability to make slight modifications on a certain feature of the image to bypass forensic models in future.

\subsection {Image Quality Assessment}
In this subsection, we show that our method is able to keep the visual quality of generated images, qualitatively and quantitatively. 

\begin{figure}[t]
\begin{center}

\includegraphics[width=0.9\linewidth]{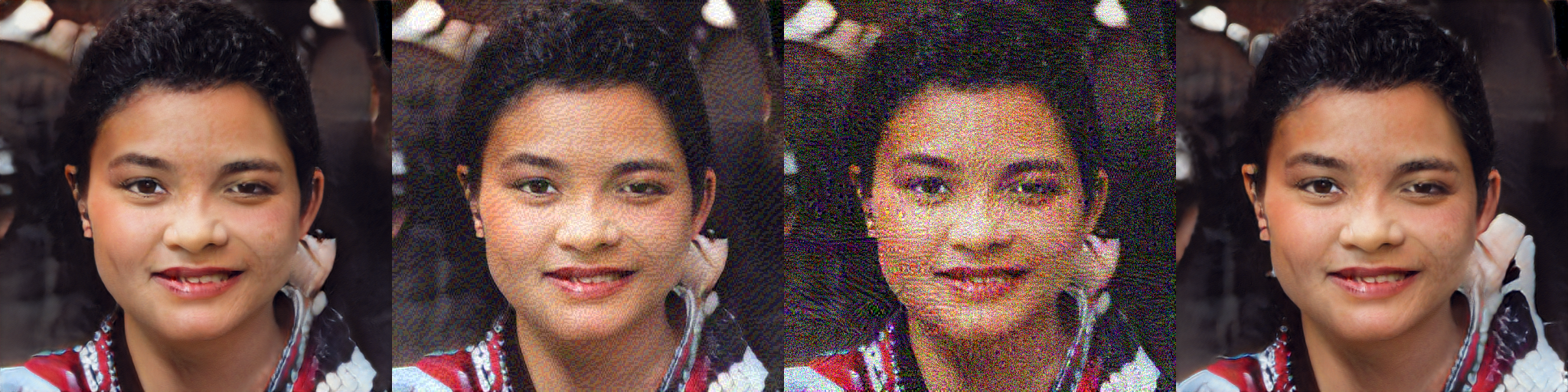}
\end{center}
   \caption{From left to right: reference image, image generated by FGSM $L_{inf}$, PGD $L_{inf}$ and proposed method, all three adversarial images can deceive the detector. Zoom in to see more details.}
\label{fig:9}
\end{figure}

Intuitively, as the total perturbation we added to latent vector has an limited range, it won't have a huge impact on the original distribution, thus the quality of the generated images can be preserved. Quantitatively, we calculate several metrics on a test dataset with 10,000 images, where reference images and images generated by three adversarial methods each account for a quarter. We also evaluate the quality of our images by user study, where 100 triplets composed of adversarial images crafted by proposed method and baseline methods are shown to 100 users, who are asked to choose the one which is most similar to the reference image for each triplet. Please refer to Table~\ref{tab:5} and Figure~\ref{fig:9} to see the result. Notice proposed method shows priority on LPIPS and user study, which better reflects human perception than other metrics.

\begin{table}[]
\begin{center}
\begin{tabular}{llll}
\hline
Metric          & Ours & PGD $L_{inf}$ & FGSM $L_{inf}$ \\ \hline
MSE ($\downarrow$) & 0.009      & 0.027     & \textbf{0.004}      \\ 
PSNR ($\uparrow$)            & 21.61      & 15.628    & \textbf{23.986}     \\
SSIM ($\uparrow$)            & 0.891      & 0.570      & \textbf{0.926}      \\ 
LPIPS  ($\downarrow$)        & \textbf{0.123}      & 1.084     & 0.507    \\
\hline
User Study    & \textbf{10000/10000} & 0/10000 & 0/10000 \\
\hline
\end{tabular}
\end{center}
\caption{Metric measuring distortion between adversarial images and reference images. proposed method has similar performance with FGSM in MSE, PSNR and SSIM, while surpassing the rest methods in LPIPS and user study by a large margin.}
\label{tab:5}
\end{table}

\section{Conclusion \& Future Work}
In this paper, we propose a novel method to generate GAN images which can bypass certain image forensic classifiers by searching on the manifold of Style-GAN. Images generated by our method can lower the forensic models' accuracy from over $90\%$ to less than $1\%$, and have better visual quality than norm-based adversarial attacks. We also explored the transferability of our images over two forensic models and deployed our method in different ensemble manners and successfully generated unrestricted adversarial images which are able to bypass several forensic models. Our method suggests more robust image forensic models are needed to identify GAN-generated fake images from real ones. In our future work, we are about to find out ways to improve transferability of our adversarial images over different model architectures. Relationship among noise level, face attributes and adversarial strength is also an attractive realm to explore.

\section*{Acknowledgements}
This work was supported by the National Key Research and Development Program of China under Grant No.2020AAA0140003, National Natural Science Foundation of China under Grant 61972395 and Grant 61772529, and Beijing Natural Science Foundation under Grant 4192058.

{\small
\bibliographystyle{ieee_fullname}
\bibliography{egbib}
}

\end{document}